\documentclass{article}

\usepackage{arxiv}

\usepackage[utf8]{inputenc} 
\usepackage[T1]{fontenc}    
\usepackage{url}            
\usepackage{booktabs}       
\usepackage{amsfonts}       
\usepackage{nicefrac}       
\usepackage{microtype}      
\usepackage{lipsum}		

\usepackage{amsmath}
\usepackage{algorithm}
\usepackage{algorithmic}
\usepackage{booktabs}
\usepackage{graphicx}

\usepackage{fancyhdr}
\title{Sparse Lifting of Dense Vectors: \\
Unifying Word and Sentence Representations}

\author{
  Wenye Li\thanks{Correspondence should be made to: Wenye Li (+86-755-84273853).} \\
  School of Science and Engineering \\
  The Chinese University of Hong Kong, Shenzhen\\
  Shenzhen, China \\
  \texttt{wyli@cuhk.edu.cn} \\
   \And
  Senyue Hao\\
  School of Science and Engineering \\
  The Chinese University of Hong Kong, Shenzhen\\
  Shenzhen, China \\
  \texttt{116010062@link.cuhk.edu.cn} \\
}

\begin{document}
\maketitle

\begin{abstract}
As the first step in automated natural language processing, representing words and sentences is of central importance and has attracted significant research attention. Different approaches, from the early {\em one-hot} and {\em bag-of-words} representation to more recent distributional dense and sparse representations, were proposed. Despite the successful results that have been achieved, such vectors tend to consist of uninterpretable components and face nontrivial challenge in both memory and computational requirement in practical applications. In this paper, we designed a novel representation model that projects dense word vectors into a higher dimensional space and favors a highly sparse and binary representation of word vectors with potentially interpretable components, while trying to maintain pairwise inner products between original vectors as much as possible. Computationally, our model is relaxed as a symmetric non-negative matrix factorization problem which admits a fast yet effective solution. In a series of empirical evaluations, the proposed model exhibited consistent improvement and high potential in practical applications.
\end{abstract}


\section{Introduction}

With tremendous theoretical and practical values, the study of natural language processing (NLP) techniques has attracted significant research attention in computer science and artificial intelligence community for many years. A key step in NLP is to represent language and text elements (words, sentences, etc.) in a form that can be processed by computers. For this task, the classical vector space model, which treats the elements as vectors of identifiers, has been routinely applied in tremendous applications for decades \cite{salton1986introduction}.

As an implementation of the vector space model, the {\em one-hot} method encodes a word by a sparse vector with exactly one element being non-zero. Accordingly, to represent a sentence, the {\em bag-of-words} scheme is naturally applied on {\em one-hot} vectors of words. Simple as it is, the representation scheme has reported good empirical results \cite{joachims1998text}. 

Besides the {\em one-hot} word representation, people have designed various representation methods, which try to encode each word into a low dimensional continuous space either as a dense vector or as a sparse one, such as the work of \cite{mikolov2013efficient,pennington2014glove,murphy2012learning,yogatama2015learning}. With these methods, a sentence vector is typically built as an average, or a concatenation, of the vectors of all sentence words. All these methods have achieved quite successful results in a variety of applications.

Despite the successful results reported, all these representation methods have some practical challenges. For example, the {\em one-hot} encoding method and the {\em bag-of-words} scheme suffer from the inherent vulnerability in measuring the similarity of words and sentences. All concatenation-based sentence representation methods need to face the expensive computational requirement in downstream applications due to the higher data dimension of each sentence. 

Motivated by both the success and the limitation of the existing encoding methods, we designed a novel word representation approach, called the {\em lifting} representation. Our method projects dense word vectors to a moderately higher dimensional space while sparsifying and binarizing the projected vectors. Intuitively, comparing with {\em one-hot} word vectors, our encoding dimension is lower yet with generally more than one non-zero elements in each vector. Comparing with other word representation schemes, our encoding dimension is typically higher while most elements are zero. In this way, the proposed approach has the potential to encode the semantics of the words without bringing about much computational burden.

Based on the proposed {\em lifting} word representation method, representing a sentence is straightforward and natural. A sentence vector can be obtained in the way of {\em bag-of-words} which sums up the {\em lifting} vectors of all sentence words. In this way, the similarity and difference of any two sentences can also be easily obtained by calculating their vector similarity or Euclidean distance.

The remainder of the paper is organized as follows. We firstly review the related work. Then we present our model in detail, and discuss the optimization issue. Next we report the empirical results, followed by conclusion and discussion.

\section{Related Work}
\label{sec:related}

\subsection{One-hot and Bag-of-Words Representation}
\label{sec:related:one-hot}

With {\em one-hot} representation model, each word is encoded as a binary vector \cite{salton1986introduction}. Only one vector element is $1$ and all other elements are $0$. The length of the vector equals to the number of words in the vocabulary. The position of the $1$ element in a vector actually gives an index of a specific word in the vocabulary.

Accordingly, a sentence can be represented easily by the bag of its words, disregarding the grammar and word order. That is, a sentence is represented as a vector of the same length with the existence of $1$s showing the existence of its words. Simple as it is, this {\em bag-of-words} representation is commonly used in document classification and information retrieval, and has reported quite successful empirical results \cite{joachims1998text}.

\subsection{Dense Representations}
\label{sec:related:dense}

It is difficult to measure the words' similarity with {\em one-hot} representation. To handle the challenges, people resorted to dense vectors. A number of methods have been designed, such as latent semantic indexing \cite{deerwester1990indexing,hofmann1999probabilistic} and Latent Dirichlet Allocation \cite{blei2003latent}. While these approaches produce a more coherent word and sentence representation than {\em one-hot} and {\em bag-of-words}, it was found that they often do not improve the empirical performance much on pattern classification tasks.

The distributional hypothesis \cite{mcdonald2001testing} attracted much attention in designing word embedding methods. Starting from a summary statistics of how often a word co-occurs with its neighbor words in a large text corpus, it is possible to project the count-statistics down to a small and dense vector for each word. These methods include latent semantic analysis \cite{dumais2004latent}, Hellinger PCA \cite{lebret2013word}, etc. More recent development along this line focuses on the prediction of a word from its neighbors in terms of dense embedding vectors. Influential work includes the classical neural language model, \cite{bengio2003neural}, the {\em word2vec} ({\em Skip-gram} and {\em CBOW}) models \cite{mikolov2013efficient}, and the {\em GloVe} algorithm \cite{pennington2014glove}. 

Different from {\em one-hot} representation, these dense vector representations are often able to quantify and categorize to some degree the semantic similarities between linguistic items based on their distributional properties in large samples of text data. This inspires a number of new applications in practice \cite{mikolov2013efficient}.

A sentence with dense word vectors cannot be represented as easily as {\em bag-of-words}. In practice, it is often processed as a concatenation of all sentence words, and therefore the sentence vector is formed by joining its dense word vectors. The concatenation-based sentence representation captures the sequential ordering of linguistic elements in the sentence with quite successful empirical results, and has been applied widely in practical applications \cite{kim2014convolutional,gehring2017convolutional}. 

However, measuring the similarity or difference between dense sentence representations is non-trivial. One cannot na\"ively apply cosine similarity or Euclidean distance for the task. Specialized distance measures, such as the {\em WMD} distance \cite{kusner2015word}, have to be designed, at the cost of significantly increased computation. Another approach is to represent a sentence by the average or weighted average of all its word vectors, and then use the averaged vectors to calculate the sentence similarities \cite{arora2017asimple}, which may achieve good results in certain applications.

\subsection{Sparse Representations}
\label{sec:related:sparse}

In addition to dense representations, recent research investigated the possibility of sparse representations. Some work starts from the co-occurrence statistics of words. Examples include the NNSE method \cite{murphy2012learning} and the hierarchical sparse encoding (FOREST) method \cite{yogatama2015learning} which learn a sparse representation of words using matrix factorization, and the sparse CBOW model \cite{sun2016sparse} which introduces an $\ell_1$ regularizer on the CBOW model to ensure the sparsity of the word vectors. 

It is also possible to start from the dense word vectors. The work of \cite{faruqui2015sparse} considers linguistically inspired dimensions as a means to induce sparsity, and proposes to transform dense word vectors to sparse, binary overcomplete word vectors through non-negative sparse coding and binarizing transformation. For the same transformation, the work of \cite{subramanian2018spine} trains a $k$-sparse autoencoder to minimize a loss function that captures the sparse constraints on word vectors. 

Comparing with dense word vectors, a key characteristic of these sparse approaches is that the resulting word vectors are found to have better interpretability than that of dense vectors. In practice, such sparse word representations have been successfully applied and achieved quite good empirical results in empirical applications such as in question answering tasks \cite{turney2017leveraging}.

Similarly to the dense word vectors, to represent a sentence with these sparse vectors, people can resort to either the concatenation-based approach or the average-based approach, depending on application scenarios.

\subsection{Dimension Expansion}
\label{sec:related:lifting}

Biological studies revealed strong evidence of dimension expansion for pattern recognition. Take the fruit fly's olfactory circuit as an example. It has $50$ Olfactory Receptor Neuron types, which are connected to $50$ projection neurons and further projected to $2,000$ Kenyon cells through sparse connections. A cell receives the firing rates from about six projection neurons and sums them up \cite{caron2013random,zheng2018complete}. Due to the strong feedback from inhibitory neuron, most cells become inactive except for the highest-firing $5\%$. These $5\%$ active cells correspond to the tag assigned to the input odor. In this way, a $50$-dimensional dense vector is mapped to a $2,000$-dimensional sparse binary vector. By simulating the fruit fly's odor detection procedure, a novel {\em fly} algorithm reported excellent performance in practical similarity retrieval applications \cite{dasgupta2017neural}.

Subsequent work along this line \cite{li2018fast} designed a {\em sparse lifting} model for dimension expansion, which is more directly related to our work. The input vectors are lifted to sparse binary vectors in a higher-dimensional space with the objective of keeping the pairwise inner product between data points as much as possible. Then the feature values are replaced by their high energy concentration locations which are further encoded in the sparse binary representation. The model reported quite good results in a number of machine learning applications.

\section{Model}
\label{sec:model}

\subsection{Sparse Lifting of Dense Vectors}
\label{sec:model:sparse}

Our work leverages recent studies on dense word representations. It starts from a word representation matrix $X\in \mathcal{R}^{N\times d}$ from either the {\em word2vec} or the {\em GloVe} representation, with which each row gives a dense vector representation of a word in a vocabulary and has been zero-centered. 

Motivated by the idea of sparse lifting, we seek a matrix $Z\in \left\{0,1\right\}^{N\times d'}$ which keeps the pairwise inner products between the row elements of $X$ as much as possible while satisfying the requirement that $\sum Z_{ij}=Nk$ where $k$ is the average number of non-zero elements specified by the user in each row vector of $Z$.

The binary constraint on the desired matrix makes the problem hard to solve. To provide a feasible solution, we resort to the following model by relaxing the binary constraint and seeking a matrix $Y\in \mathcal{R}^{N\times d'}$ to minimize the difference between $XX^{T}$ and $YY^{T}$ in the Frobenius norm:
\begin{equation}
\min_{Y}\frac{1}{2}\left\Vert XX^{T}-YY^{T}\right\Vert _{F}^{2},
\label{equ:model:spword}
\end{equation}%
subject to the element-wise constraint:
\begin{equation}
Y\ge 0.
\label{equ:model:spword:constraint}
\end{equation}

This is a symmetric non-negative matrix factorization model \cite{ding2005equivalence}. In practice, the non-negativity constraint on each element of $Y$ in Equ. (\ref{equ:model:spword:constraint}) implicitly provides some level of sparsity on the solution $Y^{*}$ \cite{lee1999learning}. When the solution $Y^{*}$ is available, we can recover the desired matrix $Z$ of {\em sparse-lifting} word vectors, or {\em lifting} vectors for short, trivially by setting $Z_{ij}=1$ if $Y^{*}_{ij}$ is among the topmost $Nk$ elements of $Y^{*}$, and setting $Z_{ij}=0$ otherwise.

\subsection{From Word to Sentence Representation}
\label{sec:model:sentence}

The {\em lifting} word representation can be easily extended to represent a sentence, in a way that is much similar to that of {\em bag-of-words}. It leads to an attribute-value representation of sentences by representing each sentence roughly as a sum of the vectors of all its words. Represented as {\em bag-of-words}, the {\em lifting} sentence representation neglects the ordering of words in a sentence, which was shown to be of less importance in a series of information retrieval tasks as reported in \cite{yang2007evaluating}.

With the {\em lifting} sentence representation, measuring the similarity or difference between two sentences becomes straightforward and trivial, which can be done just by calculating the inner product value or the Euclidean distance between the two sentence vectors.

\subsection{Differences from Previous Representations}
\label{sec:model:comparison}

The {\em lifting} representation has a number of characteristics that distinguish our proposal from existing word encoding methods fundamentally.

\textbf{Comparison with {\em one-hot} representation:} With {\em one-hot} representation, each word is represented as a sparse vector with only one element being non-zero. The word itself is treated as a non-decomposable atomic unit. The inner product of two different word vectors is always zero, i.e., each word is always {\em orthogonal} to all other words. Comparatively in each {\em lifting} word vector, the atomic unit is different. Each dimension can be potentially linked to a hidden concept, as shown in our evaluation results. A non-zero vector element is associated with the existence of a certain concept, and therefore each vector is treated as a combination of different concepts. 

\textbf{Comparison with dense representations:} In dense word vectors, the element is almost non-zero in all dimensions. It is infeasible to figure out the existence of individual concepts from dense word vectors. Furthermore, it is difficult to know the common concepts between words from their inner product values. As a result, it is difficult to identify the existence of common concepts and link them with individual dimensions.

Comparatively, when producing the {\em lifting} word vectors, the key objective is to inherit as much as possible the pairwise inner product between the original dense vectors. The inner product of two {\em lifting} vectors is zero if they share no concepts in common; or the inner product is positive to some extent if they share certain common concepts. The more concepts two words share; the higher their inner product value is. Thereby, inspecting common non-zero elements provides a simple and effective way to locate concepts shared between words.

\textbf{Comparison with existing sparse representations:} In existing sparse representations \cite{murphy2012learning,subramanian2018spine,sun2016sparse} as discussed above, the dimension of sparse word vectors is mostly equal to or very similar to the that of dense vectors. Another approach, the sparse overcomplete model \cite{faruqui2015sparse} provides a sparse representation of dense vectors in a higher-dimensional space. However, in the overcomplete model, the number of non-zero elements is comparable to the dimension of the dense vectors. For all these word representation models, it is infeasible to represent a sentence as {\em bag-of-words}.

Comparatively, our approach inherits from a sparse lifting point of view, with the explicit objective of expanding the dimension of the dense vectors while making the vector highly sparse and binary. For example, only $10$ to $20$ elements are non-zero out of $1,000$ or $2,000$ dimensions, which provides high potential to represent sentences as {\em bag-of-words}.

\subsection{Complexity Issue}

With {\em one-hot} representation, the complexity of representing a word is $O\left(1\right)$. With dense vector representations, the complexity is $O\left(d\right)$ per word where $d$ is the dimension of the word vectors. With our proposed representation, the complexity is $O\left(k\right)$ per word, assuming $k$ is the average number of non-zero elements of all $d'$-dimensional sparse vectors.

In practice, the dimension of dense vectors (i.e., the $d$) is often a few hundred. The dimension of the sparse-lifting space ($d'$) is one thousand to a few thousand. The average number of non-zero element $k$ can be from ten to less than one hundred. In our current work, the following values were used: $d=300$, $d'=1000$ and $k=20$.

To represent a sentence with $N$ words, the {\em bag-of-words} scheme needs $O\left(N\right)$ memory. For dense vector representations, when representing a sentence as an average of its word vectors, the complexity is $O\left(d\right)$; when representing a sentence as a concatenation of its word vectors, the complexity is $O\left(Nd\right)$. For our proposed representation, the worst complexity is $O\left(\min\left\{Nk,d'\right\}\right)$ per sentence. 

It can be seen that the complexity of our proposed method has a complexity that is up to a constant times higher than the {\em one-hot} representation. It is comparable to that of the averaged sentence representation of dense vectors, and significantly less than that of the concatenation-based sentence representation.

\section{Optimization Approach}
\label{sec:optimization}

\subsection{Algorithm}
\label{sec:optimization:algorithm}

The optimization model formulated in Equ. (\ref{equ:model:spword}) subject to the constraint in Equ. (\ref{equ:model:spword:constraint}) is a {\em symmetric non-negative matrix factorization} problem \cite{lee1999learning,ding2005equivalence}. Different computational approaches are possible to tackle the problem \cite{lee2001algorithms}. We resort to a simple relaxation approach:
\begin{equation}
\min_{W,H\ge 0}\left\|XX^{T}-WH^{T}\right\|^2_F+\alpha\left\|W-H\right\|^2_F.
\label{equ:nonsym:model}
\end{equation}
Here we seek two matrices $W$, $H$ of size $N\times d'$, and $\alpha>0$ is a scalar parameter for the trade-off between the approximation error and the difference of $W$ and $H$. With the relaxation, we force the separation of the unknown matrix $Y$ by associating it with two different matrices $W$ and $H$. Given a sufficiently large value of $\alpha$, the matrix difference dominates the objective value and the solutions of $W$ and $H$ will tend to be close enough so that the word vectors will not be affected whether $W$ or $H$ are used as the result of $Y$.

The key to solving the problem in Equ. (\ref{equ:nonsym:model}) is by solving the following two {\em non-negative least squares} (NLS) sub-problems \cite{kuang2015symnmf}:
\begin{equation}
  \min_{W\ge 0}
  \left\|
  \begin{matrix}
  H\\
  \sqrt{\alpha}I_{d'}
  \end{matrix}
  W^T-
  \begin{matrix}
  XX^{T}\\
  \sqrt{\alpha}H^T
  \end{matrix}
  \right\|_F^2,
  \label{equ:nonsym:sub1}
\end{equation}
and
\begin{equation}
  \min_{H\ge 0}
  \left\|
  \begin{matrix}
  W\\
  \sqrt{\alpha}I_{d'}
  \end{matrix}
  H^T-
  \begin{matrix}
  XX^{T}\\
  \sqrt{\alpha}W^T
  \end{matrix}
  \right\|_F^2,
  \label{equ:nonsym:sub2}
\end{equation}
where $I_{d'}$ is the $d'\times d'$ identity matrix. Solving the sub-problems in the two equations in an iterative way will lead to a stationary point solution, as long as an optimal solution is returned for every NLS sub-problem encountered.

In practice, this alternating algorithm can be further simplified by successively solving only one sub-problem only. The details are shown in Algorithm \ref{alg:anls}.

\begin{algorithm}
\caption{An Alternating NLS Algorithm for:
$min_{W,H\ge 0}\left\|XX^{T}-WH^T\right\|_F^2
+\alpha\left\|W-H\right\|_F^2$}
\label{alg:anls}
\begin{algorithmic}
  \STATE INPUT: $X$, $d'$
  \STATE Initialize $H_{0}$
  \STATE $i\leftarrow 0$
  \REPEAT
  \STATE $W_{i+1}\leftarrow H_{i}$
  \STATE $H_{i+1}\leftarrow \arg\min_{H\ge 0}  \left\|
    \begin{matrix} W_{i+1}\\
    \sqrt{\alpha}I_{d'}\end{matrix} H^T-
    \begin{matrix} XX^{T}\\
    \sqrt{\alpha}W^T_{i+1}\end{matrix}
    \right\|_F^2$
  \STATE $i\leftarrow i+1$
  \UNTIL convergence
  \STATE OUTPUT: $W_i$, $H_i$
\end{algorithmic}
\end{algorithm}

The key step in the alternating NLS algorithm is to solve the NLS sub-problem, which is a classical problem arising in many applications and can be solved efficiently via the rather standard routines such as the active-set method or the projected gradient method \cite{kim2008nonnegative}. A detailed discussion of these NLS algorithms goes beyond the scope of this paper and is therefore omitted.

For choosing the parameter $\alpha$, we can start from a small positive value, for example $1$, and gradually increase its value to a very large number. The increase is terminated when the ratio of $\left\|W-H\right\|_F/\left\|H\right\|_F$ is negligible.

\subsection{Scalability: An ADMM Perspective}

In Algorithm \ref{alg:anls}, the key step is to solve an NLS sub-problem for $H_{i+1}$. By using the project gradient method \cite{kim2008nonnegative}, the complexity of the NLS sub-problem is $O\left(Nd'^{2}\right)$, which works efficiently to moderately large problems (say $N=100,000$, $k=1,000$) on a mainstream workstation. 

For even larger un-structured problems, the computation and memory requirements grow rapidly and may become infeasible. As a remedy, we can re-write the problem in Equ. (\ref{equ:nonsym:model}) by seeking matrices $W,H\ge 0$ to minimize:
\begin{equation}
\left\Vert XX^{T}-WH^{T}\right\Vert^{2}_{F}+\left\Vert\Gamma^{T}\left(W-H\right)\right\Vert^{2}_{F}+\frac{\rho}{2}\left\Vert W-H\right\Vert^{2}_{F}
\end{equation}
where $\Gamma$ denotes the matrix of dual variables, and $\rho$ is a positive number. This becomes a problem that can be approximately solved by the alternating direction method of multipliers (ADMM) \cite{boyd2011distributed}. ADMM executes efficiently in parallel and distributed computing platforms with high scalability, and provides a potential solution to very large optimization problems. A detailed discussion of the ADMM method goes beyond the scope of this paper and is therefore omitted.

\section{Evaluations}
\label{sec:evaluations}

\subsection{General Settings}
\label{sec:evaluations:settings}

To evaluate the performance of the proposed word representation method, we carried out a series of evaluations. Our {\em lifting} vectors were generated from the dense word vectors released by the authors of {\em word2vec} and {\em GloVe}.
\begin{itemize}
  \item {\em CBOW}: 300-dimensional word2vec vectors trained on about 100 billion tokens \footnote{\url{https://code.google.com/archive/p/word2vec/}}.
  \item {\em GloVe}: 300-dimensional word vectors trained on about 6 billion tokens \footnote{\url{https://nlp.stanford.edu/projects/glove/}}.
\end{itemize}

We trained the {\em lifting} vectors with the $50,000$ most frequent words out of the {\em CBOW} and {\em GloVe} word vectors respectively. The expanded dimension of the trained vectors were set to $d'=1,000$. After training, on average $20$ elements of each vector were set non-zero, i.e. the hash length $k=20$. The results reported in this paper are just based on this setting. Besides, we have also varied different combinations of the parameters within the range of $d'=1,000/2,000/5,000$ and $k=10/20$. The evaluation results are quite similar and are therefore omitted.

In the evaluation, six benchmark datasets were used.
\begin{itemize}
  \item CUSTREV: A collection of $3,774$ customer reviews of various products. On average, each sentence has $15$ words. The task is to predict positive or negative reviews \cite{hu2004mining}.
  \item MPQA: A collection of $10,606$ news articles. On average, each sentence has $3$ words. The task is to predict positive or negative opinion polarities \cite{wiebe2005annotating}.
  \item RT-POLARITY: A collection of $10,662$ movie review sentences. On average, each sentence has $18$ words. The task is to classify a review as being positive or negative \cite{pang2005seeing}.
  \item STSA-binary: An extension of the RT-POLARITY dataset with $8,741$ sentences. On average, each sentence has $19$ words \cite{socher2013recursive}.
  \item Subjectivity: A collection of $10,000$ sentences. On average, each sentence has $20$ words. The task is to classify subjective or objective sentences \cite{pang2005seeing}.
  \item TREC: A collection of $5,692$ TREC questions. On average, each sentence $8$ words. The task is to classify a question into six question types \cite{li2002learning}.
\end{itemize}

Besides the {\em one-hot}, {\em CBOW} and {\em GloVe} representations, our {\em lifting} vectors were compared with the following representations:
\begin{itemize}
  \item {\em FOREST}: $52$-dimensional word vectors based on hierarchical sparse coding of the co-occurrence statistics between words and contexts \footnote{\url{http://www.cs.cmu.edu/~ark/dyogatam/wordvecs/}}.
  \item {\em NNSE}: $300$-dimensional word vectors by decomposing topical relatedness information through the sparse SVD method \footnote{\url{http://www.cs.cmu.edu/~bmurphy/NNSE/}}.
  \item {\em OVERCOMPLETE}: $1,000$-dimensional sparse overcomplete word vectors generated from the {\em CBOW} and {\em GloVe} vectors respectively \footnote{\url{https://github.com/mfaruqui/sparse-coding/}}.
\end{itemize}

For {\em one-hot} and {\em lifting} methods, each sentence vector is represented as {\em bag-of-words}. For all other word vectors, we represent each sentence as an average of fifty word vectors \cite{arora2017asimple}. For a small number of sentences that have more than fifty words, the extra words were cut off. For most sentences that have less than fifty words, the corresponding elements were padded with zeros. In addition to the the averaged representation, a concatenation-based representation which treats each sentence vector as a concatenation of {\em CBOW} vectors was also included in the experiment, combined with the Word Mover Distance ({\em WMD}) to measure sentence similarities \cite{kusner2015word}.

\subsection{Interpretability}
\label{sec:evaluations:interpretability}

We firstly carried out a small experiment to test the potential of the {\em lifting} method on representing the semantics of words. Similarly to other sparse word vectors, the dimensions of the {\em lifting} vectors are found to be more interpretable than those of dense word vectors. As discussed above, a key assumption underlying our proposal is to regard each word as a collection of hidden concepts. The concepts are potentially interpretable, and are reflected directly through individual dimensions. Therefore the dimension can be associated with physical meanings.

The experimental results shown in Table \ref{tab:semantics} support our assumption. In the $1,000$-dimensional {\em lifting} word vectors generated from the {\em CBOW} vectors, after manual inspection we found that one dimension is closely related to the concept of countries. Its value is $1$ for most country words, and is $0$ for other words. Similar and evident patterns were also found for color words, flower words and words of IT companies, etc.

\begin{table}[t]
  \caption{Examples of {\em lifting} vector components that are closely associated with words' semantics after sparse-lifting of {\em CBOW} vectors. Column 1: one dimension that is related to countries; Column 2: one dimension that is related to colors; Column 3: one dimension that is related to flowers; Column 4: one dimension that is related to IT companies.}
  \label{tab:semantics}
  \centering
  \begin{tabular}{|@{}p{1.8cm}@{}|@{}p{1.8cm}@{}|@{}p{1.8cm}@{}|@{}p{1.8cm}@{}|}
    \toprule
    \cmidrule{1-4}
    {\em DIM-1} & {\em DIM-2} & {\em DIM-3} & {\em DIM-4}  \\
    \midrule
    {\em america} & {\em black}   & {\em daisy}   & {\em amazon} \\
    {\em brazil}  & {\em blue}  & {\em heather} & {\em cisco} \\
    {\em canada}  & {\em golden}  & {\em iris}  & {\em dell} \\
    {\em china}   & {\em green}   & {\em jasmine} & {\em google} \\
    {\em france}  & {\em grey}  & {\em lilac}   & {\em hp} \\
    {\em germany}   & {\em orange}  & {\em lily}  & {\em ibm} \\
    {\em india}   & {\em pink}  & {\em rose}  & {\em intel} \\
    {\em japan}   & {\em purple}  & {\em sunflower}   & {\em lenovo} \\
    {\em kingdom}   & {\em white}   & {\em tulip}   & {\em microsoft} \\
    {\em russia}  & {\em yellow}  & {\em violet}  & {\em oracle} \\
    {\em ...}     & {\em ...}   & {\em ...}   & {\em ...} \\
    \bottomrule
  \end{tabular}
\end{table}

\subsection{Sentiment Analysis}
\label{sec:evaluations:sentiment}

\begin{figure*}[!t]
\centering
\includegraphics[width=0.48\textwidth,height=2.0in]{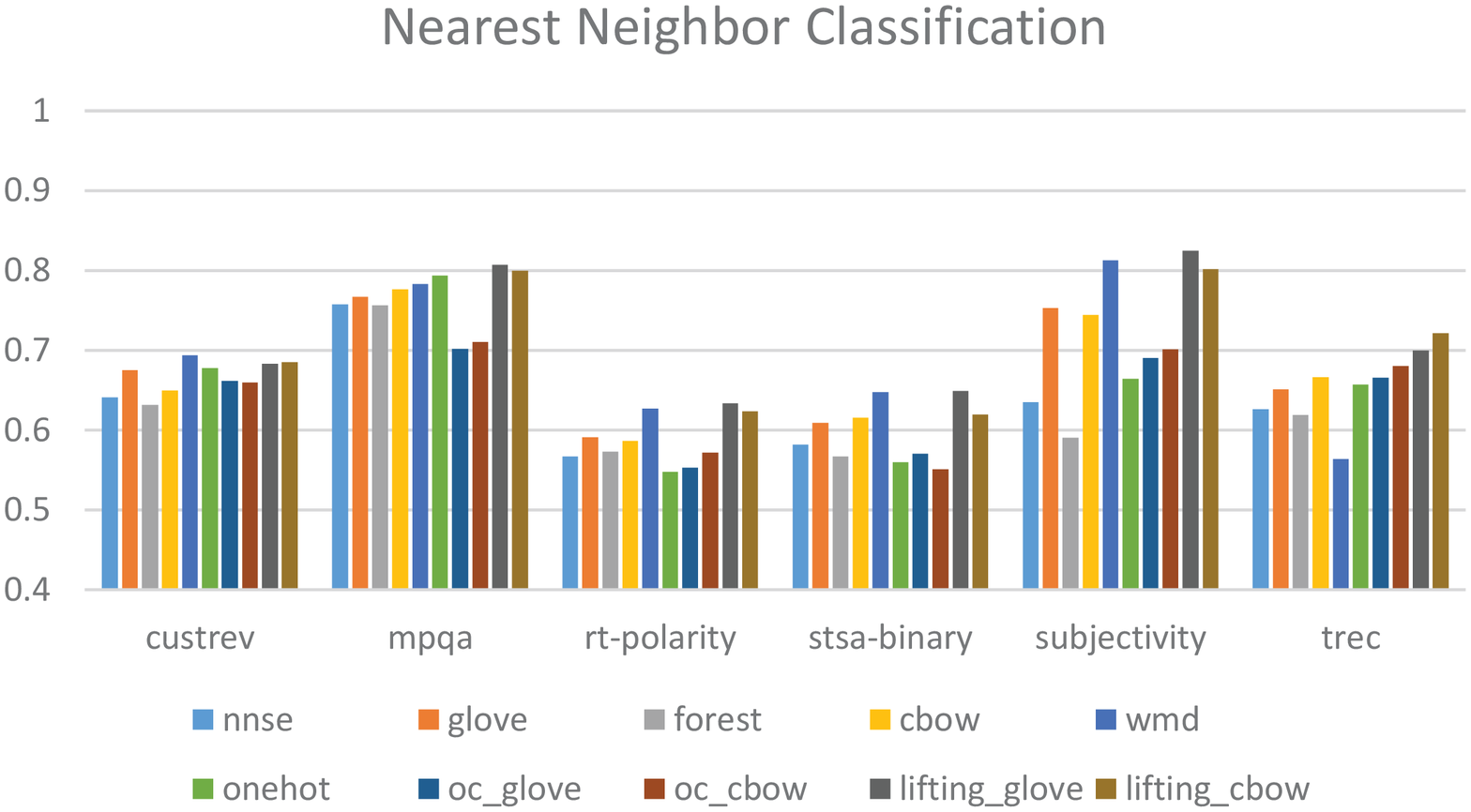}
\includegraphics[width=0.48\textwidth,height=2.0in]{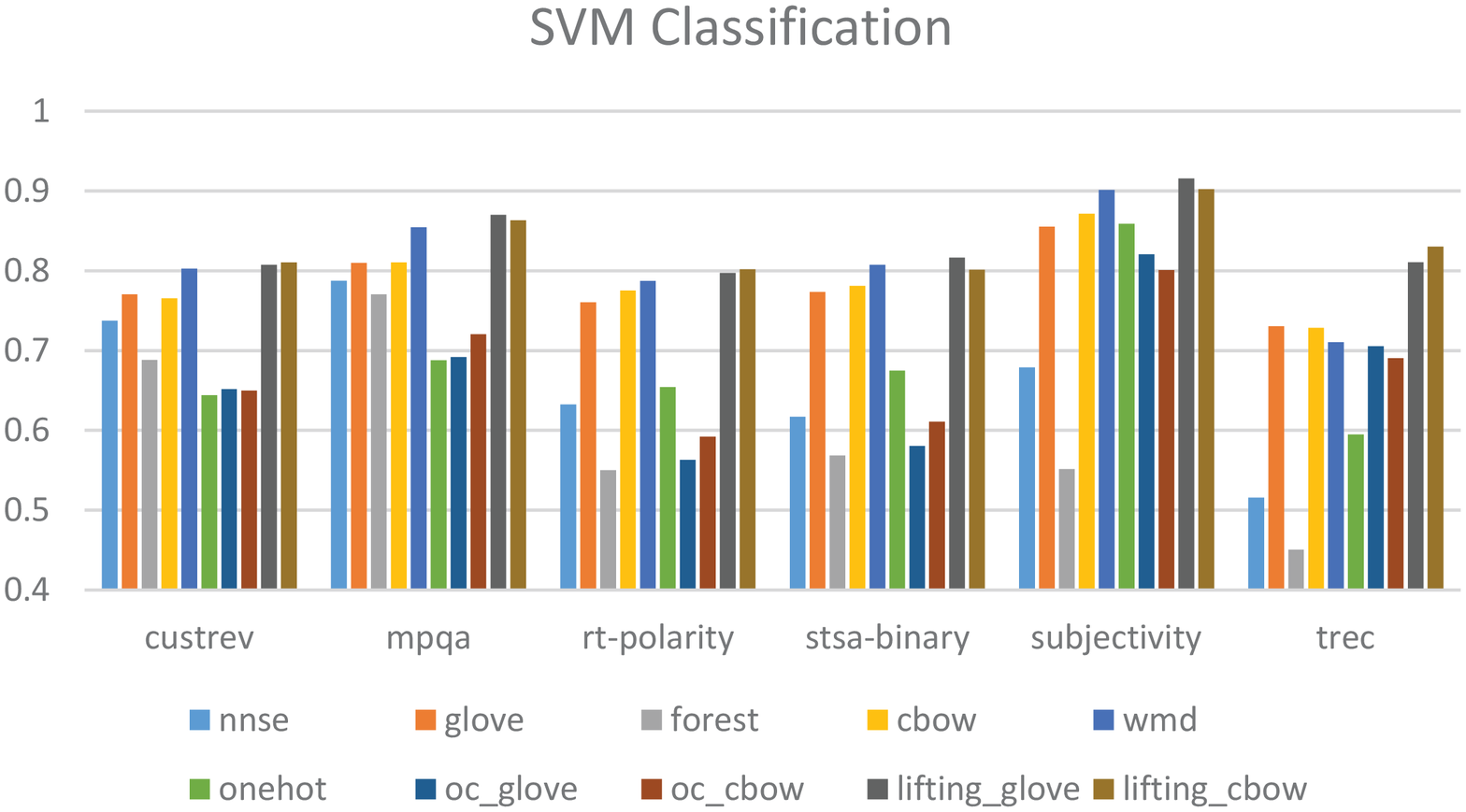}
\caption{Sentiment classification accuracies with ten-fold cross validation on different representations and distance measures. Horizontal: different datasets; Vertical: classification accuracies. Left: Nearest Neighbors Classification; Right: Support Vector Machines Classification.} 
\label{fig:sentiment}
\end{figure*}

\begin{figure*}[!t]
\centering
\includegraphics[width=0.48\textwidth,height=2.0in]{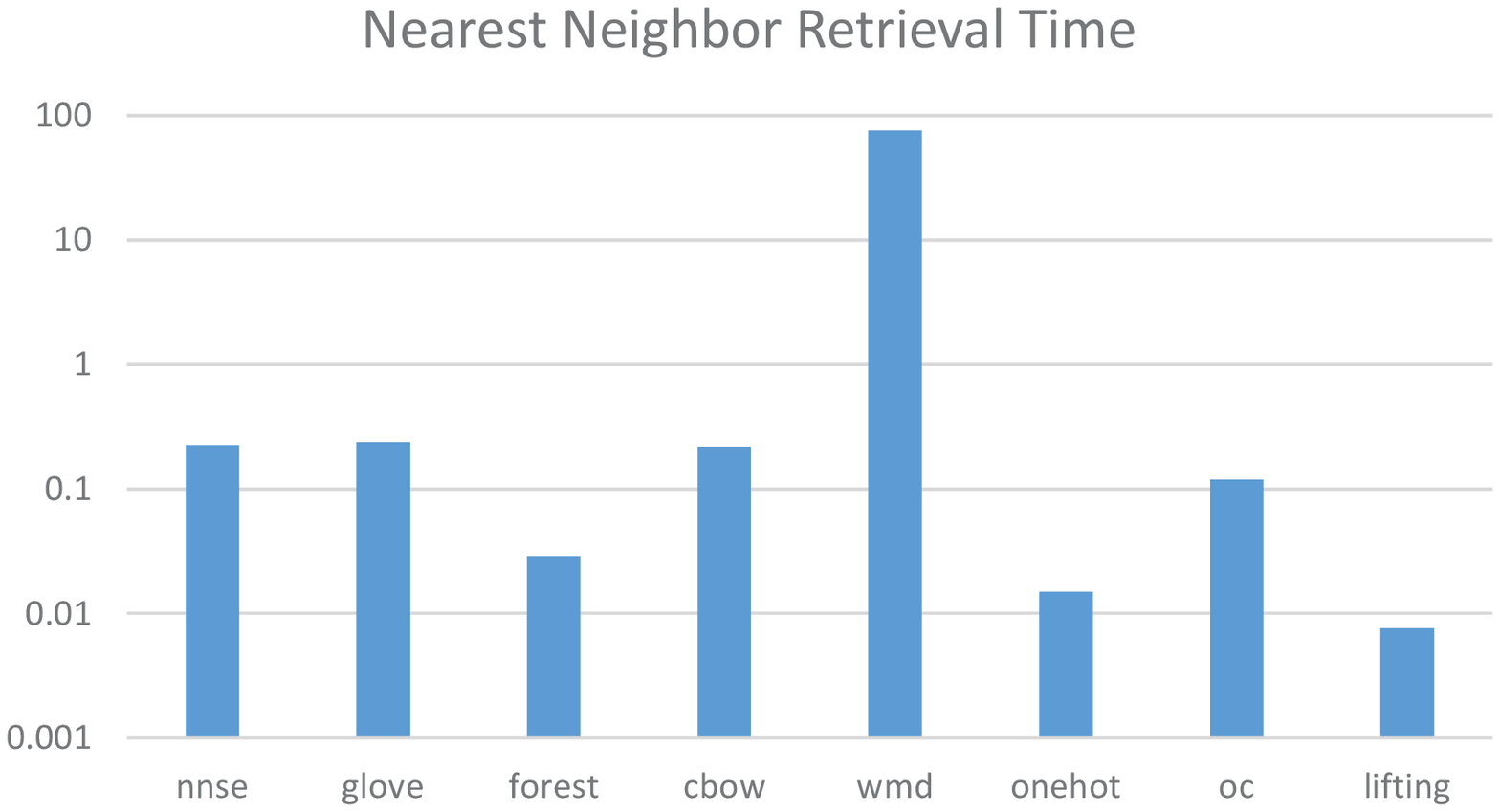}
\includegraphics[width=0.48\textwidth,height=2.0in]{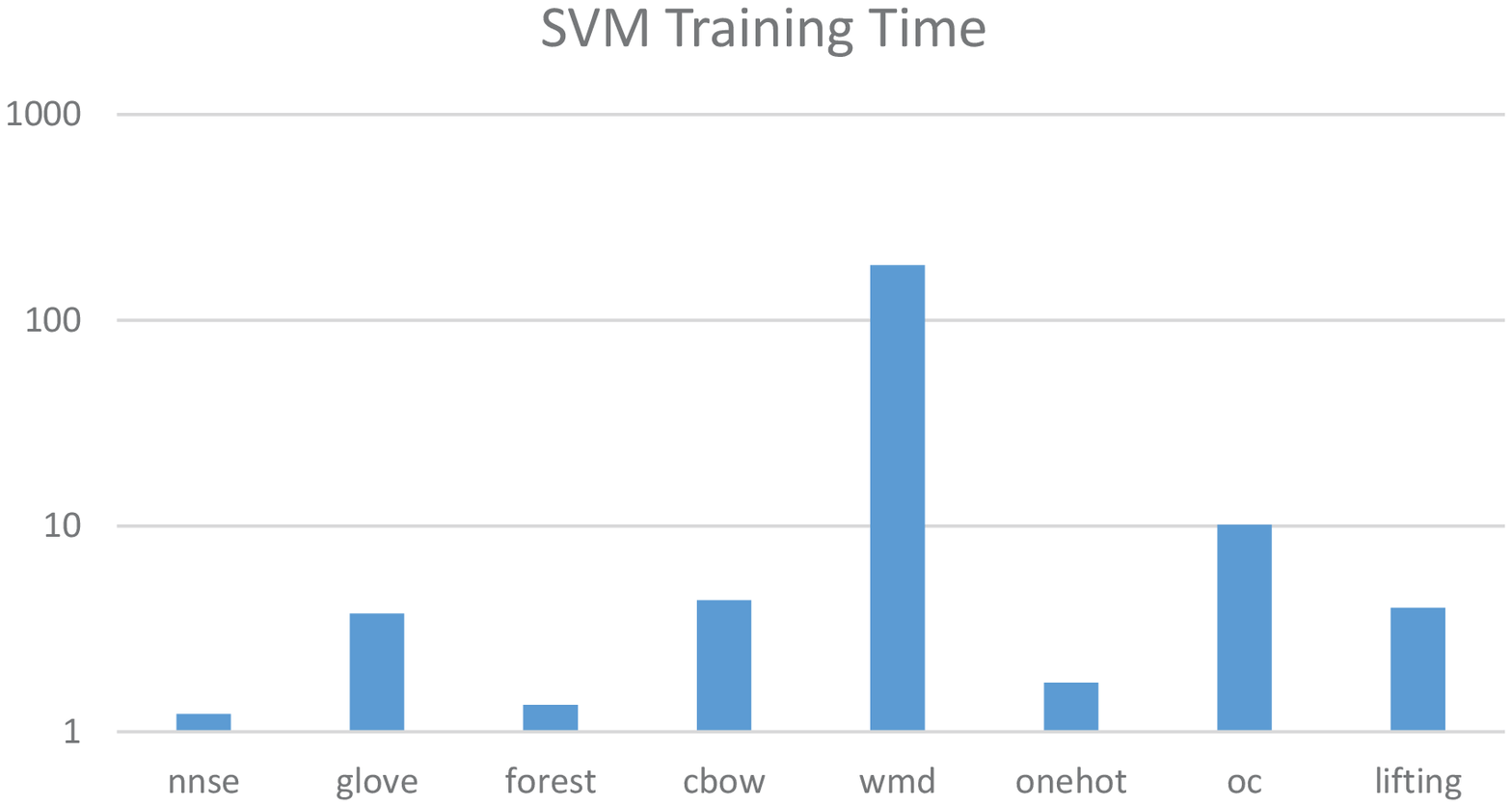}
\caption{Running time of nearest neighbors classification and support vector machines classification on CUSTREV dataset with $90\%$ of samples as training data and the rest $10\%$ as testing data. Horizontal: different representations; Vertical: the running time (seconds) in log-scale. Left: Nearest Neighbors Classification; Right: Support Vector Machines Classification with Gaussian kernel and default parameters.} 
\label{fig:speed}
\end{figure*}

We investigated the performance of the sparse-lifting vectors on sentiment analysis applications on the six benchmark datasets. On each dataset, we trained a classifier with different representations of sentences by the {\em nearest neighbors} (NN) algorithm and the {\em support vector machines} (SVM) algorithm respectively. For the SVM classification, we used the Gaussian radial basis function kernel, with the default kernel and cost parameters \cite{chang2011libsvm}.

The comparison was made against with a number of representation schemes, including {\em NNSE}, {\em GloVe}, {\em FOREST}, {\em CBOW}, and {\em OVERCOMPLETE} (denoted by {\em oc\_cbow} and {\em oc\_glove} in the figure) representations with Euclidean distance measure. The classification accuracies are depicted in Fig. \ref{fig:sentiment}. Each result in the table is an average accuracy of $10$-fold cross validations.

The {\em one-hot/bag-of-words} representation, simple as it is, reported acceptable results in these distance-based classification applications when working with the SVM algorithm; but its results on NN classification were not as good. This result is consistent with the previously reported studies in the work of \cite{joachims1998text}. The {\em CBOW} and concatenation-based sentence vectors, when being combined with {\em WMD} to measure the sentence similarity, reported quite good accuracies on both NN and SVM classifications, which shows the necessity of using a specialized sentence distance measure for the concatenation-based representation. The performances of the averaged sentence representation with {\em Glove/CBOW/NNSE/FOREST} vectors, however, seemed not quite satisfactory. The classification results on some datasets in NN classification were only slightly better than the random guess, which left much room to improve.

Comparatively, our proposed representations (denoted by {\em lifting\_glove} and {\em lifting\_word2vec} respectively in the figure), brought improved results in the experiments. Compared with {\em CBOW} and concatenation-based sentence vectors with the {\em WMD} measure, our proposed representations reported comparable (if not better) classification accuracies on most datasets; while on the TREC dataset which has six categories, both of our representations reported much better results. The results verified again the potential of representing word and sentence semantics by the proposed {\em lifting} encoding method.

\subsection{Running Speed}
\label{sec:evaluations:speed}

As a practical concern, we further recorded the running time on NN classification and SVM classification with different representations. We recorded the query time of the NN classifier with $90\%$ of samples used as training samples and the rest $10\%$ used as testing samples on CUSTREV dataset, which needs to compute the distances between each pair of testing and training samples. 

The experiment was performed in a computer server with $44$ CPU cores enabled for parallel execution. The results are depicted in Fig. \ref{fig:speed}. From the results we can see that with highly sparse and binary representations, the query time of the {\em lifting} representation and of the {\em one-hot/BOW} representation reported significantly superior results over other methods on NN classification. With other representations, the computation involves operations on real numbers, it needs magnitudes more time to retrieve the nearest neighbor for a given query. This becomes even worse when the WMD method is used to measure the distance, which can be thousands times slower than our proposed sparse and binary representation.

Besides the NN classification, we also recorded the running time of training an SVM classifier with different representations by using the libSVM package \cite{chang2011libsvm} with $90\%$ of samples on CUSTREV dataset used as the training set. With a Gaussian kernel and the default setting of parameters, it took less than $10$ seconds to train an SVM classifier with the {\em lifting} vectors and {\em bag-of-words} sentence representation. Similar results were found on SVM training with the {\em Glove}, {\em CBOW} and {\em OVERCOMPLETE} vectors and averaged sentence representations. All these results are tens of times faster than training with the {\em CBOW} vectors and concatenation-based sentence representation in {\em WMD} distance, which took around $300$ seconds.

The SVM training with our proposed {\em lifting} representation run slower than training with the {\em NNSE}, {\em FOREST} and {\em one-hot} representations, which took around $2$ to $4$ seconds. This is due to the lower dimension of {\em NNSE} vectors ($300$ dimensions) and {\em FOREST} vectors ($52$ dimensions), or even sparser representation of {\em one-hot}. Considering the significantly improved classification accuracies brought by the proposed {\em lifting} method, this should be a reasonable computational overhead.

In addition to the Gaussian kernel and the default parameter setting, we also repeated the experiment with the linear kernel and the polynomial kernel with different parameter settings, which reported much similar trends in running time. The results are therefore omitted here.

\section{Conclusion}
\label{sec:conclusion}

Motivated by the recent advances in biological studies, our work designed a novel sparse lifting word representation method which projects given dense word vectors into a higher dimensional space while ensuring the sparsity and binarization of the projected vectors. The word representation is simple, flexible and can be naturally extended to represent sentences in a simple way of of {\em bag-of-words}.

Comparing with existing popular word and sentence vector representations, our proposed sparse-lifting representation has been shown to be an appropriate representation for distance-based learning tasks and has reported significantly improved results in sentiment analysis tasks. The improvement provides us with high confidence to apply the method in wider practical applications. Meanwhile, many modeling and algorithmic issues still remain to be studied for the proposed approach, as promising as it appears to be. To deepen understanding, further work will be necessary to investigate the relationships between the proposed word embedding approach and existing methodologies.

The proposed method can be regarded as a lightweight encoding method. It represents a sentence directly from its word vectors with trivial or even no additional computation at all. Comparatively, much recent work also studied heavyweight encoding methods. A highly influential heavyweight method is the {\em BERT} model \cite{devlin2018bert}, which reported excellent results in many language processing tasks. However, such models typically need a large training set and involve heavy computation in training, which could potentially limit their applications. Comparatively, we believe the lightweight methods, such as the {\em lifting} model, have much less practical limitation, which could yield more general applications and can be applied where heavyweight methods are not applicable, such as in sentiment analysis tasks with limited samples, as shown in our paper.

For the future work, one particular interesting point is to investigate the possibility of applying the proposed {\em lifting} representation with the deep learning models \cite{goodfellow2016deep}. Popular neural network models, such as recurrent neural networks and convolutional neural networks, are not often used together with sparse representations. Accordingly, designing new deep network architectures for sparse data may become necessary.

Another interesting point for the future work is about the relationship between the embedding dimension ($d'$) and the hash length ($k$). Our current work mainly resides on the biological evidence and the practical experience, and suggests a highly sparse vector (i.e., $k\ll d'$) for word representation. However, it would be desirable to seek an optimal value of $k$ (in certain sense), which remains open for further study. With the recent achievement in related areas \cite{yin2018dimensionality}, we strong believe the direction could be feasible for investigation.

\section{Acknowledgments}

This work was supported by Shenzhen Fundamental Research Fund (JCYJ20170306141038939, KQJSCX20170728162302784).



\end{document}